\documentstyle[eaclap]{article}

\title{\vspace{-0.5in}
Implementation and evaluation of a German HMM for POS disambiguation}

\author{Helmut Feldweg\\
Department of Linguistics\\
University of T\"ubingen\\
Wilhelmstr.\ 113\\
72074 T\"ubingen\\
Germany\\
Helmut.Feldweg@uni-tuebingen.de}

\begin{document}

\maketitle
\vspace{-0.5in}

\begin{abstract}

A German language model for the Xerox HMM tagger is presented.  This
model's performance is compared with two other German taggers with
partial parameter re--estimation and full adaption of parameters from
pre--tagged corpora. The ambiguity types resolved by this model are
analysed and compared to ambiguity types of English and French.
Finally, the model's error types are described. I argue that although
the overall performance of these models for German is comparable to
results for English and French, a more exact analysis demonstrates
important differences in the types of disambiguation involved for German.

\end{abstract}

\section{Background}

Since the late '80s part--of--speech (POS) disambiguation using Hidden
Markov Models (HMM) has been a widespread method for tagging texts. Despite
this fact, little work has been done so far toward employing this
technology for the disambiguation of German texts (cf.\ \cite{Wothke1993},
\cite{Schmid1994}). Earlier work of the author (Feldweg 1993 and
1995)\nocite{Feldweg1993a}\nocite{Feldweg1995b} within the framework
of a project on corpus based development of lexical knowledge bases
(ELWIS) has produced LIKELY, a straightforward implementation of the
Viterbi algorithm employing an HMM whose parameters were obtained from
a pre--tagged text corpus.  Since then the original tag set was
redefined, making the tagged corpus used to train the LIKELY tagger
obsolete.
\smallskip

Within a current project on adapting bilingual dictionaries for online
comprehension assistance (COMPASS, LRE 62--080), the need arose for a
POS--disambiguator to facilitate a context sensitive dictionary
look--up system.  As the COMPASS project makes ample use of Xerox
technology for its core look--up engine and for POS disambiguation for
languages other than German, the obvious thing to do was to develop a
German language model for the Xerox tagger.
\medskip

The following section describes the implementation of this new
model. In section~\ref{sec:Comparison} the results obtained using the
new model are compared with the results from previous models. An
analysis of the types of disambiguation involved in these models
is presented in section~\ref{sec:AmbiguityTypes}. The model's error
types are analysed in section~\ref{sec:ErrorTypes}, and conclusions
are drawn in section ~\ref{sec:Conclusion}.

\section{Implementation of the German language model}
\label{sec:Implementation}

The version of the Xerox tagger used for the implementation described
here is the DDS Tagger version 1.1 \cite{Kupiec1994}. This version
differs from the current version (1.2) of the Xerox Tagger as
described in
\cite{Cutting1992} in that the DDS Tagger accommodates lexicons and
class guessers in the form of external finite--state transducers.
Implementing a new language model for this tagger involves supplying:

\begin{enumerate}
\renewcommand{\labelenumi}{(\arabic{enumi})}
\item  a definition of the tag set to be used by the HMM,
\item  a lexicon listing word forms with their equivalence classes, that is
       the list of POS tags that can be assigned to the word form,
\item  a class guesser that assigns equivalence classes to words not
       covered by the lexicon,
\item  a set of initial transition biases,
\item  a set of initial symbol biases,
\item  a sufficiently large text for training the HMM,
\item  a reference text with correctly assigned POS tags,
\item  a tokenizer that recognizes words in free text.
\end{enumerate}

The following paragraphs describe these components in more detail.
\medskip

(1)\qquad
The tag set used in the implementation is the smaller version of the
two tag sets developed jointly by the Universities of Stuttgart and
T\"ubingen, referred to as the ELWIS tag set (cf.~\cite{Thielen1994a}
and \cite{Thielen1995a}). It consists of a total of 42 POS tags plus
three tags for punctuation and a special tag for truncated words. The
tags used in the ELWIS tag set are given in table~\ref{tab:ELWIS-Tagset}.
\smallskip

\begin{table}[htbp]
\begin{center}
%\begin{list}{}{\labelwidth1.5cm
%	\itemindent0cm \labelsep 0.2cm \leftmargin1.7cm
%	\rightmargin0cm \itemsep0cm}
\begin{tabular}{|l|p{5.2cm}|}\hline
Label	&	 Part--of--Speech \\\hline
NN	&	 noun \\
NE	&	 proper noun \\
VFIN	&	 finite verb \\
VINF	&	 infinitive verb\\
VIZU	&	 infinitive verb with {\it zu\/}\\
VPP	&	 past participle\\
ART	&	 article\\
ADJA	&	 adjective, attributive \\
ADJD	&	 adjective, adverbial\\
PPER	&	 personal pronoun, irreflexive\\
PPERRF	&	 personal pronoun, reflexive and irreflexive\\
PRF	&	 pronoun, reflexive\\
PPOSS	&	 pronoun, possessive, substantive\\
PPOSAT	&	 pronoun, possessive, attributional\\
PROS	&	 pronoun, demonstrative and indefinite\\
PROAT	&	 pronominal adverb\\
PWS	&	 pronoun, interrogative, substantive\\
PWAT	&	 pronoun, interrogative, attributive\\
PRELS	&	 pronoun, relative, substantive\\
PRELAT	&	 pronoun, relative, attributive\\
PNFL	&	 pronoun, non inflecting\\
PALL	&	 pronoun, forms of {\it all--\/}\\
PBEID	&	 pronoun, forms of {\it beid--\/}\\
PVIEL	&	 non inflecting, non attributive forms of {\it viel--\/},
		 {\it wenig\/} etc.\\
CARD	&	 numbers, cardinal\\
ADV	&	 adverbs\\
KOUI	&	 conjunction, subordinating with infinitive completion\\
KOUS	&	 conjunction, subordinating with finite completion\\
KON	&	 conjunction, coordinating\\
KOKOM	&	 conjunction, comparative\\
APPR	&	 preposition\\
APPO	&	 postposition\\
APZR	&	 circumposition, right part\\
ITJ	&	 interjection\\
PTKZU	&	 particle, infinitival {\it zu\/}\\
PTKNEG	&	 particle, negation\\
PTKVZS	&	 separated verbal prefix\\
PTKANT	&	 particle, answer\\
PTKA	&	 particle, adjectival or adverbial\\
TRUNC	&	 truncated word\\
\$.	&	 punctuation, sentence delimiting\\
\$,	&	 punctuation, phrase delimiting\\
\$(	&	 punctuation, other\\\hline
\end{tabular}
\medskip

Detailed guidelines on the use of the individual tags are available in
\cite{Thielen1994b}.
\caption{The ELWIS tag set}
\label{tab:ELWIS-Tagset}
\end{center}
\end{table}

(2)\qquad
For each word form the lexicon gives the set of POS tags that can be
assigned to that word. This set may consist of one (for unambiguous
words) or more tags and is referred to as the word's equivalence
class.  The lexicon used in this implementation was derived from
Lingsoft's GERTWOL morphological analyzer, which uses finite state
transducers, by mapping the morpho--syntactic labels generated by
GERTWOL onto the ELWIS tag set. The lexicon is realized as a finite
state transducer for the DDS Tagger. Alternative mapping rules were
developed to generate a lexicon with ELWIS tags from the German
lexical database of the Centre for Lexical Information
\cite{CELEX1993}.
\smallskip

(3)\qquad Class guessers for the Xerox tagger assign potential POS
tags to unknown words according to a surface analysis of the word
form. In addition to the common practice of mapping POS tags according
to the words' suffixes, this implementation makes use of the case of
the initial letter of a word --- which is highly significant for POS
assignment in German.  The class guesser also takes care of
POS--assignment for abbreviations, special symbol sequences and
language external material in the text.
The class guesser, like the lexicon, is a  finite state transducer.

\smallskip

(4)\qquad
The model is trained using a set of initial transition
biases, including both positive and negative constraints on tag
sequences.  Although the model can be trained without initial biases,
the performance of the resulting model increases significantly if
appropriate initial biases are used.

The biases in the model consisted for the most part in specifications
of the most plausible successor tags for each tag in the tag set.
They were constructed manually and refined in a series of subsequent
training and evaluation runs.
\smallskip

(5)\qquad
Initial symbol biases are an additional set of biases used to define
preferences for tag assignment given a particular equivalence
class. Only a very few symbol biases were defined before evaluation of
the first training runs, mainly to reflect biases towards equivalence
classes used in the class guesser. The majority of symbol biases were
added to correct misguided biases chosen during the training
processes.
\smallskip

(6)\qquad
The two texts used for training the HMM were selected from the German
data contained on the ECI's Multilingual CD--ROM \cite{ECI1994}: a
200,000 and 2,000,000 word sample from Summer 1992 issues of the
German newspaper {\em Frankfurter Rundschau}.
\smallskip

(7)\qquad
The reference texts were also taken from the
{\em Frankfurter Rundschau}, but do not overlap with the training
texts. The reference texts amount to a total of approximately 20,000
running words, which were manually tagged and checked.
\smallskip

(8)\qquad
The current version of the implementation uses a
straightforward tokenizer accepting one line per token.  Training
texts are pretokenized using an external tokenizer written in lex.

\subsection{Performance}

The best results of this implementation were obtained running 20
iterations of training over the 200,000 word training text, using a
total of 50 transition and 17 symbol biases. With this configuration
of training parameters, the resulting HMM assigned $3.33\;\%$
incorrect tags when run on the reference texts and compared with the
manually assigned tags.

\section{Comparison with other German models}
\label{sec:Comparison}

The main advantage of the Xerox tagger when compared with earlier
implementations of HMM taggers is that it can be trained using
untagged text. However, the performance of the resulting HMM is very
poor if no initial biases are used to help the training process find
suitable parameters.

For comparison, the evaluation procedure used to evaluate the
implementation of the HMM tagger described in the preceding section
was repeated without using any of the initial biases. The result was a
poor performance of the resulting HMM with an error rate of
$14.11\;\%$.

Choosing initial biases to help train a model is a subtle task in that
it not only requires sound knowledge of the tag set used and the
target language the model is aiming at, but it also requires a
``feel'' for how the initial biases may be modified during a given
number of training iterations. It is also sometimes frustrating that
the linguistic knowledge used to create the initial biases gets
``optimized'' or ``trained'' away in subsequent iterations of
training.

To overcome these disadvantages, hybrid technologies have been
developed that combine free text training methods with parameter
estimation from pre--tagged texts. In such a setting, initial
transition and symbol biases are replaced by frequencies of tag
sequences and tag instantiation from a relatively small pre-tagged
corpus. The counted frequencies are taken as an approximation to the
model's probabilities and get smoothed in a small number
of training iterations.

In order to see what could be gained for a German language model by such
a hybrid technology, I used extensions to the Xerox tagger
developed at the University of Stuttgart that facilitate
initialization of an HMM with values obtained from a pre--tagged
corpus (cf.\ \cite{Schmid1994}). Initial
parameters were obtained by counting transition and symbol frequencies
in a manually tagged 24,000 word corpus taken from the newspaper
{\em Stuttgarter Zeitung}, that was kindly made available by the
University of Stuttgart. These initial parameters were adjusted in a
single training iteration using Xerox's Baum--Welch implementation for
parameter re--estimation. The texts used for training and as a
reference for evaluation were the same as the ones used in the
implementation described in section~\ref{sec:Implementation}. The
resulting HMM had an error rate of $3.14\;\%$.  The superior
performance of this model confirms results presented by
\cite{Briscoe1994}, \cite{Merialdo1994},
and \cite{Elworthy1994} for English: empirically obtained initial
values for transition and output probabilities with a small number of
training iterations lead to significantly better results than
intuitively generated biases do.

On the other end of the scale of parameter production for HMM POS
disambiguators is the extraction of parameters exclusively on the
basis of (larger) pre--tagged text corpora, with no Baum--Welch
re--estimation involved. Such an implementation is described in earlier
work of the author (cf.\ \cite{Feldweg1993a} and
\cite{Feldweg1995b}). For this model error rates of
$3.16\;\%-7.29\;\%$ (depending on the coverage of the underlying
lexicon) were reported. These results, however, are not directly
comparable to the implementations described in this paper, since the
tag sets employed differ to some extent.

\section{Assessment of ambiguity types}
\label{sec:AmbiguityTypes}

In the preceding sections the evaluation of the model was purely quantitative.
Performance was measured as the percent of mismatches between the output
generated by the HMM and the tags assigned by manual tagging. Although
this error rate is an appropriate measure for the performance of an
HMM given a particular reference text, it says little about the amount
of disambiguation done by the tagger, and nothing about the
ambiguity types that were involved in the disambiguation process.

The difficulty of disambiguation can be quantified by the ambiguity
rate: the number of possible tag assignments divided by the number of
words in a given text. The test text used to evaluate the German model
described in section
\ref{sec:Implementation} has an ambiguity rate of $1.51$, this is,
the lexicon provides an average of $1.51$ tags for each word in the
text.

\begin{table*}[htb]
\begin{center}\small
\begin{tabular}{*{3}{|l|p{3.64cm}|}}\hline
\multicolumn{2}{|c||}{\normalsize\phantom{\large D}English\phantom{\large D}} &
\multicolumn{2}{|c||}{\normalsize French} &
\multicolumn{2}{|c|}{\normalsize German} \\ % \hline
f(ec) & elements of equiv.\ class & f(ec) & elements of equiv.\ class
& f(ec) & elements of equiv.\ class \\\hline
.0701 & NN VB & .0862 & DET-PL PREP & .0772 & ART PROS PRELS \\
.0441 & VBD VBN & .0678 & DET-SG PC & .0265 & PTKVZS APPR \\
.0357 & JJ NN & .0263 & NOUN-SG VERB-P1P2 VERB-P3SG & .0255 & NE NN \\
.0301 & NNS VBZ & .0233 & ADJ-SG NOUN-SG & .0252 & VINF VFIN \\
.0224 & AT NP & .0174 & ADJ-SG PAP-SG & .0119 & ADV KON \\
.0118 & JJ NN VB & .0158 & DET-PL PC & .0117 & ART PROS PROAT CARD \\
.0111 & IN RB & .0125 & PC PREP & .0116 & VPP ADJD \\
.0097 & PPO PPS & .0119 & DET-SG NUM PRON & .0095 & VPP ADJD VFIN \\
.0097 & IN JJ RB & .0118 & DET-SG PREP & .0089 & PROS PROAT \\
.0094 & CS DT WPS & .0109 & ADJ-PL NOUN-PL & .0086 & PTKVZS APPO APPR
APZR \\\hline
\end{tabular}\smallskip\\
f(ec) = relative frequency of equivalence class
\smallskip

\caption{Elements of 10 most frequent ambiguous equivalence classes
	for English, German and French}
\label{tab:EquivalenceClassesFreq}
\end{center}
\end{table*}

In an effort to compare the German model with what is reported for
other languages, English and French tagged texts were
analysed. Both texts contained approximately $10,000$ words
and were tagged using an English resp.\ French language model for the
Xerox tagger.

The tag set used to annotate the English text is a slightly modified
version of the Brown tag set, consisting of a total of 72 tags. The
ambiguity rate for this text is $1.41$.  For the French text the tag
set described in
\cite{Chanod1994} with 37 different tags is used. Here the ambiguity
rate is $1.52$.  In terms of ambiguity rates, the English, French, and
German texts are thus quite comparable.

In order to compare the types of ambiguities that had to be resolved
by the different language models for the HMM tagger, relative
frequency tables of equivalence classes were computed for each of the
three texts. The top most frequent equivalence classes for the three
languages are listed in table \ref{tab:EquivalenceClassesFreq} together
with their relative frequencies.

If the table is viewed in terms of the major word classes of noun,
verb, adjective, adverb, and closed--class forms, the following
predominant ambiguity classes for English can be distinguished:

\begin{itemize}
\item noun vs.\ verb ({\it share, offer, plan\/}),
\item adjective vs.\ noun ({\it public, million, high\/}),
\item closed--class vs.\ noun ({\it a\/}),
\item adjective vs.\ noun vs.\ verb ({\it return, field\/}),
\item closed--class vs.\ adverb ({\it by, about, below\/}),
\item and closed--class vs.\ adjective vs.\ adverb ({\it round, next\/}),
\end{itemize}
For French the major ambiguity types are:
\begin{itemize}
\item noun vs.\ verb ({\it affaire, bout, place\/}),
\item adjective vs.\ noun ({\it demi, moyen, responsable\/},
\item adjective vs.\ verb ({\it appliqu\'e, devenu, fabriqu\'e\/}),
\item closed--class vs.\ adjective (numeral {\it un\/}).
\end{itemize}
The elements of the most frequent ambiguity types for German, however,
belong to the same major word classes, with only a few exception such as:
\begin{itemize}
\item closed--class vs.\ adjective (numeral {\it einen, einer\/}),
\item and verb vs.\ adjective ({\it fehlgeschlagen, bekannt\/}),
\end{itemize}
with the latter reflecting a subtle distinction in the
German tag set (VPP vs.\ ADJD: participle as modifier vs.\
non--attributive adjectives).

The comparison of the most frequent ambiguity types shows a
significant difference between the German model on the one hand and
the English and French models on the other. In German most of the
effort is going into subclassification within major word classes,
while in English and French a good deal of disambiguation work is
devoted to separate major word classes.

\section{Assessment of error types}
\label{sec:ErrorTypes}

The differences in ambiguity types of the models also have effects on the
types of errors produced by the German model. Again, errors mainly
affect the assignment of words to subclasses within one major word
class.

Table \ref{tab:ErrAssess} shows the most common errors produced by the
German model. The entries are sorted by decreasing frequencies
relative to the total number of mismatches between the manually and
automatically tagged texts.  The first column gives the relative
frequenciy and the second column lists the tag chosen by the German HMM
tagger. In the second column, the number following the slash indicates the
number of elements in the equivalence class from which the model had
to choose. A missing number indicates that there was only one
choice in the lexicon.  The third column show the ``correct'' tag, as
chosen by the human tagger.

\begin{table}[htb]
\begin{center}
\begin{tabular}{|r|l|l|}\hline
Rel.Freq & HMM & Human \\\hline
$0.0900$ & VINF/2	&	VFIN\\
$0.0790$ & NN/2	&	NE\\
$0.0648$ & NE/2	&	NN\\
$0.0521$ & NN	&	NE\\
$0.0332$ & NE/7	&	NN\\
$0.0316$ & VPP/3	&	VFIN\\
$0.0269$ & VPP/3	&	ADJD\\
$0.0269$ & ADV/2	&	KON\\
$0.0237$ & APPO/6	&	APPR\\
$0.0205$ & PROS/3	&	ART\\
$0.0205$ & PROS/2	&	PWS\\
$0.0158$ & PWAV/4	&	KOKOM\\
$0.0158$ & PRELS/3	&	PROS\\
$0.0158$ & ART/3	&	PROS\\
$0.0158$ & ART/3	&	PRELS\\
$0.0142$ & VFIN/2	&	VINF\\
$0.0126$ & VPP/2	&	ADJD\\
$0.0126$ & VINF/3	&	VFIN\\
$0.0126$ & KOUS/2	&	APPR\\
$0.0126$ & KON/4	&	KOKOM\\\hline
\end{tabular}
\medskip
\caption{20 most common error types of German HMM}
\label{tab:ErrAssess}
\end{center}
\end{table}

The most common (accumulated) error is the confusion of proper nouns
and common nouns --- a result of the fact that both proper nouns and
common nouns are both capitalized in German.  The fourth line of
table~\ref{tab:ErrAssess} represents a special case of this error for
which the HMM model can not be held responsible: no ambiguity was
indicated in the lexicon, so the model had nothing to choose
from. This occurs most frequently when common nouns are used as proper
nouns (e.g.\ {\it in die gehobene Mittelklasse plaziert Renault ab 6.\
M\"arz den \underline{Safrane}\/}), where one would not expect to add a
tag ``proper noun'' for every noun in the lexicon.

The second most frequent error type involves confusion of infinitives
and 1st and 3rd pers.\ pl.\ finite present tense forms.  These are
homographs in German that are notoriously hard to disambiguate within
a narrow context.

The difficulty of distinguishing between non--attributive, adjectival
usage of participles (i.e.\ {\it er ist geladen\/}) and participles
proper (i.e.\ {\it er hat den Wagen geladen\/}) was mentioned in the
preceding section. In addition a number of these forms may also be
used as finite verbs (i.e.\ {\it erhalten, geh\"ort\/}), and this is a
further source of errors.

Almost all of the remaining errors are misassignments within closed
classes, including well--known errors due to long distance phenomena,
such as those resulting from the confusion of relative pronouns,
demonstrative pronouns and articles in sentences like: {\it
\underline{die} einmal f\"ur die Buchproduktion erfa{\ss}ten
\underline{Texte}\/} or: {\it doch \underline{der} wollte nicht,
\underline{das} falle auf.\/}

\section{Conclusion}
\label{sec:Conclusion}

Despite the hypothesis that the free word order of German leads to
poor performance of low order HMM taggers when compared with a
language like English, the overall results for German are very much
along the lines of comparable implementations for English, if not
better.  It can be argued that the disadvantage of free word order for
HMM taggers is compensated for by richer morphology and the
additional disambiguation cue of having upper and lower case initial
letters to distinguish POS membership. The latter, however, greatly
hinders the recognition of proper nouns, the most common type of
error, responsible for approximately $20\;\%$ of the model's
$3.33\;\%$ mistakes.
\smallskip

It is important to notice that the types of disambiguation carried out
by the tagger for German are significantly different from the
disambiguation work for English and French. While in English and
French a fair number of disambiguations involve separating major POS
classes such as verb, noun, and adjective, most of the work performed
in the German model involves disambiguation between subclasses of one
main category, such as finite vs.\ infinitive verb, noun vs.\ proper
noun, different sub-categories of pronouns, etc.
\smallskip

This finding has consequences for the COMPASS project, where POS
disambiguation is employed as one means of disambiguating word senses
to facilitate precise dictionary look--up. While this technique helps
to confine word senses for English and French, it is of little help for
word sense disambiguation in German.

However, the German model was useful for the project because a tagged
reference corpus was required for lexicographic work in order to adapt
existing bilingual dictionaries.  The tagger was used to annotate all
of the 50 million word German corpora contained on the ECI
Multilingual Corpus 1 CD--ROM.
\bigskip

\section*{Acknowledgements}

I would like to thank Helmut Schmid of the University of Stuttgart for
providing extensions of parameter initialization for the Xerox
Tagger and Jean--Pierre Chanod and Lauri Karttunen of the Rank Xerox
Research Laboratory Grenoble for making available the English and
French tagged texts and lexicons.  I would also like to acknowledge
valuable advice from Tracy Holloway King and Steven Abney, who
commented on earlier versions of this paper.

This work has been supported in part by the Ministry for Science and
Research of the Land Baden--W\"urttemberg under the project ``Corpus
Based Development of Lexical Knowledge Bases (ELWIS)'' and by the
Commission of the European Community under the LRE project ``Adapting
Bilingual Dictionaries for Online Assistance (COMPASS, LRE 62--080)''.

%% \bibliographystyle{acl}
%% \bibliography{/home/elwis/doc/homemade/bibl/bibl}

\end{document}